\documentclass[letterpaper, 10 pt, conference]{ieeeconf}  

\IEEEoverridecommandlockouts                              

\usepackage[dvipsnames]{xcolor}
\usepackage{lipsum}
\usepackage{graphics} 
\usepackage{subfigure}
\usepackage{epsfig} 
\usepackage{times} 
\usepackage{amsmath} 
\usepackage{amssymb}  
\usepackage{cite}

\makeatletter
\let\NAT@parse\undefined
\makeatother

\usepackage[pagebackref,breaklinks=true,colorlinks,bookmarks=false]{hyperref}
\usepackage[accsupp]{axessibility}  
\hypersetup{
linkcolor=BrickRed
,citecolor=Green
,filecolor=Mulberry
,urlcolor=NavyBlue
,menucolor=BrickRed
,runcolor=Mulberry
,linkbordercolor=BrickRed
,citebordercolor=Green
,filebordercolor=Mulberry
,urlbordercolor=NavyBlue
,menubordercolor=BrickRed
,runbordercolor=Mulberry
}
\usepackage{cleveref}

\usepackage{booktabs}       
\usepackage{colortbl}
\newcommand{\tbr}[1]{\textbf{\textcolor{BrickRed}{#1}}}
\newcommand{\tbg}[1]{\textbf{\textcolor{Green}{#1}}}
\newcommand{\tbb}[1]{\textbf{\textcolor{NavyBlue}{#1}}}

\title{\LARGE \bf
LiDAR-based 4D Occupancy Completion and Forecasting
}
\author{Xinhao Liu$^{1,*}$, Moonjun Gong$^{1,*}$, Qi Fang$^2$, Haoyu Xie$^1$, Yiming Li$^1$, Hang Zhao$^3$, Chen Feng$^{1,\dagger}$
\\
\thanks{$^1$New York University. $^2$University of Toronto. $^3$Tsinghua University.}%
\thanks{$^*$Equal Contribution.}%
\thanks{$^\dagger$Corresponding author. Email: \texttt{cfeng@nyu.edu}. The work was supported by NSF 2238968 and 2345139 grants; and in part through the NYU IT High Performance Computing resources, services, and staff expertise.}%
}

\begin{document}

\maketitle

\thispagestyle{plain}
\pagestyle{plain}

\begin{abstract}

Scene completion and forecasting are two popular perception problems in research for mobile agents like autonomous vehicles. Existing approaches treat the two problems in isolation, resulting in a separate perception of the two aspects. In this paper, we introduce a novel LiDAR perception task of \textit{Occupancy Completion and Forecasting} (OCF) in the context of autonomous driving to unify these aspects into a cohesive framework. This task requires new algorithms to address three challenges altogether: (1) sparse-to-dense reconstruction, (2) partial-to-complete hallucination, and (3) 3D-to-4D prediction. To enable supervision and evaluation, we curate a large-scale dataset termed \textit{OCFBench} from public autonomous driving datasets. We analyze the performance of closely related existing baselines and variants on our dataset. We envision that this research will inspire and call for further investigation in this evolving and crucial area of 4D perception. Our code for data curation and baseline implementation is available at \url{https://github.com/ai4ce/Occ4cast}.

\end{abstract}

\section{INTRODUCTION}
A comprehensive understanding of the dynamic 3D environment is crucial for mobile agents such as autonomous vehicles. While already widely used in mobile robotics (e.g., localization and mapping), grid-centric representation of the 3D environment, particularly occupancy grids represented as 3D voxels, has also gained prominence in perception tasks in autonomous driving. Compared to alternative representations, grid-centric representation has advantages in terms of fine-grained comprehension, efficient sensor fusion, robustness to occlusion, and flexibility in planning~\cite{shi2023grid}. 

\begin{figure}[ht]
    \centering
    \includegraphics[width=0.95\linewidth]{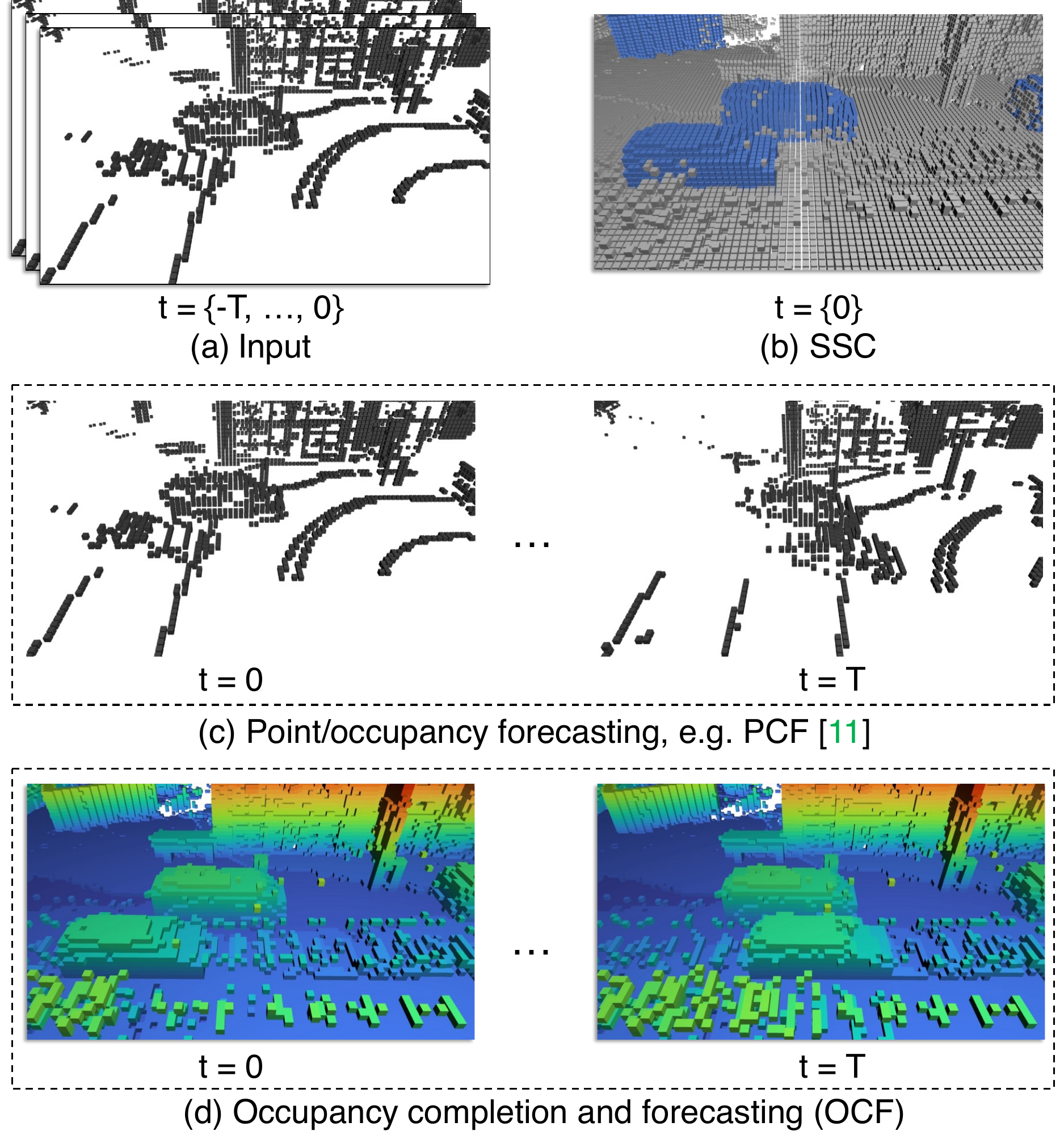}
    \caption{\textbf{Distinctions between OCF and related tasks}. (a) All tasks take a sequence or a single LiDAR sweep as input. (b) SSC aims to densify, complete, and semantically predict on the $t=0$ frame. (c) Point/occupancy forecasting outputs a sparse and Lagrangian specification of the scene geometry's motion field. (d) OCF combines scene completion and occupancy forecasting in a spatial-temporal way, outputting a dense and Eulerian motion field. The color gradient in (d) indicates the z-coordinate.}
    \label{fig:related_task}
    \vspace{-4mm}
\end{figure}

\begin{figure*}[ht]
    \centering
    \includegraphics[width=0.95\linewidth]{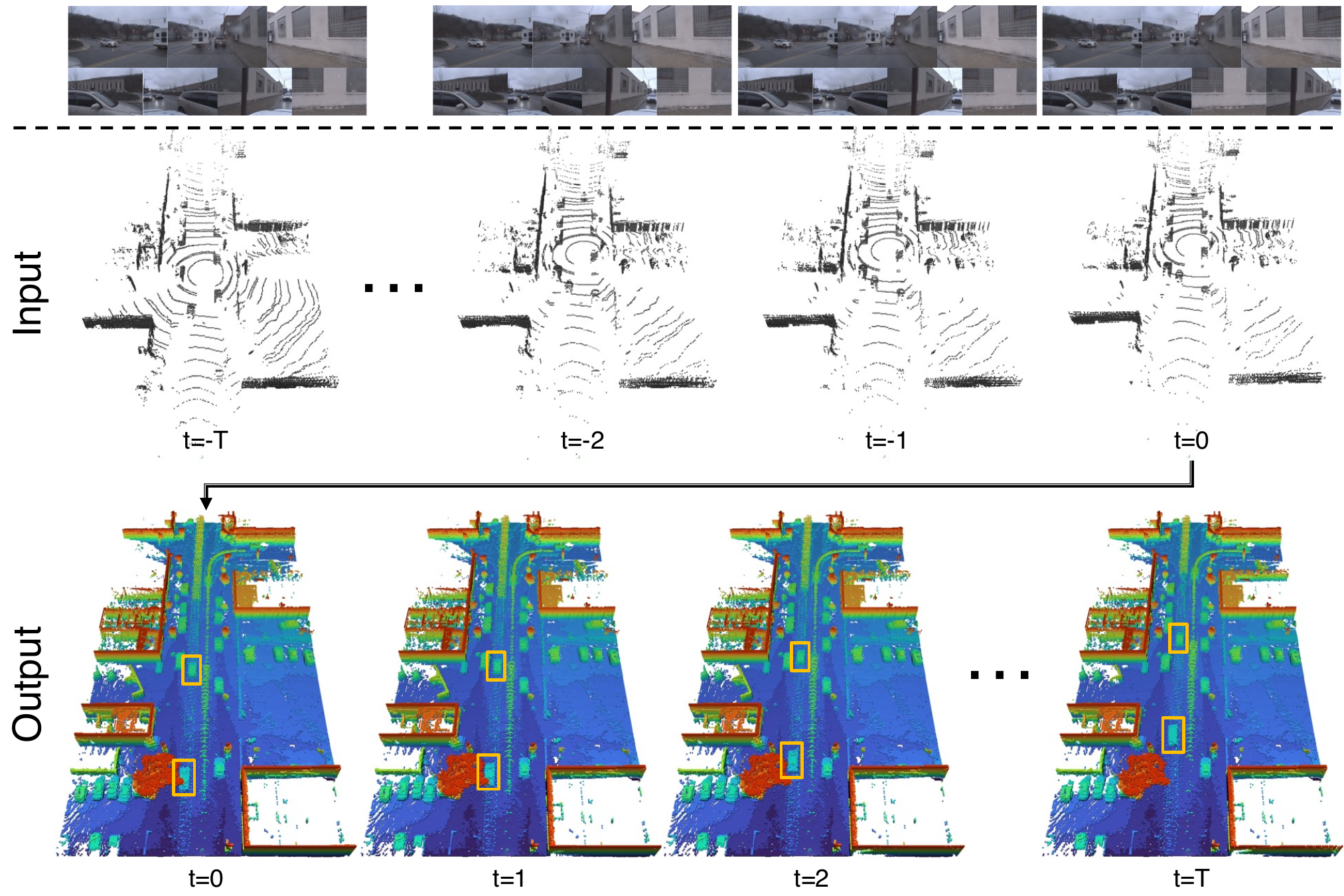}
    \caption{\textbf{Illustration of the OCF task}. The input is provided as a sequence of sparse LiDAR sweeps from $t=-T$ to $t=0$. The output is a sequence of densified and completed voxels from $t=0$ to $t=T$. The color gradient indicates the z-coordinate of each voxel. All point clouds and voxels are expressed in the coordinate frame at $t=0$. The \textcolor{YellowOrange}{yellow} bounding boxes highlight typical moving objects. The images on the top row are only for visualization and are not included in the input. Figure best viewed in color.}
    \label{fig:ocf}
    \vspace{-5mm}
\end{figure*}

Multiple perception tasks have emerged based on occupancy voxel grids~\cite{li2022bevformer,hu2021fiery,zhang2022beverse,hu2023planning}. Among these, semantic scene completion (SSC)~\cite{song2017semantic,roldao2020lmscnet,cao2022monoscene,li2023voxformer} and occupancy forecasting~\cite{khurana2022differentiable,khurana2023point} have gained increasing attention. SSC enables segmentation of occupancy grids in both visible and occluded spaces within the environment, while occupancy forecasting focuses on predicting the temporal change of the scene. Both tasks share the common objective to hallucinate either the \textit{spatially} or \textit{temporally} unseen environment from partial observation. 

While existing methodologies have historically treated occupancy completion and forecasting as separate tasks, it is evident that the two tasks are inherently interconnected. Analogously to human perception, where the observation of the front of a vehicle allows for the simultaneous mental reconstruction of its rear and the prediction of its future state, perception algorithms must possess the capacity to concurrently address completion and forecasting. By doing so, mobile agents can capture the intricacy among objects, entities, and dynamics, thereby elevating their awareness of the evolving environment.

With this motivation, our goal is to enable models to learn scene completion and occupancy forecasting concurrently. In this paper, we introduce the task as \textbf{occupancy completion and forecasting} (OCF). As depicted in \Cref{fig:ocf}, given a series of consecutive LiDAR sweeps, the model is expected to produce a sequence of completed occupancy grids represented as voxels. \Cref{fig:related_task} shows that OCF distinguishes from SSC in that it additionally incorporates temporal forecasting, and from point/occupancy forecasting in that a completed dense occupancy is predicted. Subsequent to this formulation are several challenges inherent to the task:

(1) \textit{Sparse-to-dense reconstruction}: LiDAR sensors inherently produce sparse data, with increasing sparsity as the distance from the sensor grows. Addressing this challenge requires algorithms to effectively interpolate sensor data gaps and implicitly reconstruct scene surfaces. While object-level solutions have been extensively investigated~\cite{fei2022comprehensive}, this challenge is only addressed tangentially within SSC.

(2) \textit{Partial-to-complete hallucination}: Another facet of scene completion in SSC involves the hallucination of occluded voxels from observable ones. It requires algorithms to reason both visible and invisible spaces while maintaining spatial coherence. Although this is possible to achieve by utilizing prior knowledge~\cite{wang2020point}, it is impractical to maintain a knowledge base for complex and dynamic autonomous driving scenarios with multiple objects and structures.

(3) \textit{3D-to-4D prediction}: Occupancy forecasting extends the challenge by predicting the evolution of the completed voxels over time. This transition from 3D to 4D demands algorithms capable of modeling the temporal dynamics of the environment, accounting for the movement of objects and entities within it. Unlike Lagrangian specification used in occupancy flow or forecasting works\cite{Reza2022occupancy,khurana2023point}, achieving accurate forecasting on dense voxel grids employs Eulerian specification, which naturally requires a more comprehensive understanding of the dynamics in the environment.

To address these challenges, an immediate question arises: how to establish the ground truth for supervision and evaluation? Previous work~\cite{khurana2023point} suggests using point clouds as a proxy for occupancy. Although point cloud data is readily available from public autonomous driving datasets, it introduces potential bias to the supervision process due to depth rendering errors~\cite{khurana2023point}. On the contrary, we follow SSC literature~\cite{behley2019semantickitti,tian2023occ3d,li2023sscbench} to generate ground truth by aggregating and voxelizing multiple LiDAR sweeps. Additionally, to mitigate egomotion, we incorporate coordinate unification that allows perception algorithms to focus on modeling the evolution of the environment only.

To elucidate the intricacies of the newly introduced OCF task and to investigate effective methodologies for addressing its challenges, we conducted a comprehensive benchmarking study. Through these experiments, we sought to establish a baseline for future research. We also show these structures help improve the model performance of the most related existing work~\cite{khurana2023point}.

Our contributions are summarized as the following:
\begin{itemize}
    \item We propose the OCF task, which mandates a spatial-temporal dense 4D perception from sparse 3D inputs.
    \item We generate a large-scale dataset termed \textit{OCFBench} by leveraging public autonomous driving data.
    \item We benchmark existing baselines for the OCF task and provide insights for future research in this problem.
\end{itemize}

\section{RELATED WORK}

\textbf{Scene completion.} Scene completion originally emerged in the context of indoor scene reconstruction~\cite{zheng2013beyond,ramakrishnan2018sidekick,song2018im2pano3d,yang2019extreme,popovic2021volumetric}. SSCNet~\cite{song2017semantic} elevate the problem by taking semantics into consideration. Recently, SSC in autonomous driving has gained growing interest and can be categorized by input modality into camera-based~\cite{cao2022monoscene,li2023voxformer,huang2023tri,zhang2023occformer,wei2023surroundocc} and LiDAR-based~\cite{song2017semantic,roldao2020lmscnet,cheng2021s3cnet,xia2023scpnet} approaches. While these methods achieve a satisfactory understanding of the environment in both semantic and geometric aspects, they do not account for the temporal evolution of the scene. Furthermore, existing SSC methods heavily rely on human-annotated point-level semantic labels, which is prohibitive and restricts the training scalability. In contrast, our proposed OCF task mitigates this barrier by focusing solely on the geometric perception of the environment and additionally extends the task to include temporal forecasting.

\textbf{Point cloud forecasting.} It naturally serves as a self-supervised learning task due to the easy access to sequential point cloud data~\cite{sun2020novel,deng2020temporal,zhang2021cloudlstm}. Recent literature also proposes to use it as a proxy problem for motion forecasting~\cite{weng2022s2net,ye2021tpcn,mersch2022self}, pose estimation~\cite{weng2021inverting}, and occupancy forecasting~\cite{khurana2023point} in autonomous driving scenarios. However, \cite{khurana2023point} argues that point cloud forecasting forces the algorithms to learn sensor intrinsics and extrinsics, making it less ideal for autonomous systems. OCF builds upon this idea by adopting a grid-centric representation of the environment with all input and output frames unified to the same coordinate system.

\textbf{Occupancy forecasting.} The problem originates from forecasting semantic occupancy grids on bird's-eye view (BEV)~\cite{schreiber2019long,casas2021mp3,sadat2020perceive,Reza2022occupancy}. \cite{khurana2022differentiable} extends the problem to be independent of expensive semantic labeling by limiting the problem to geometric perception. The most relevant work to our paper in this thread is~\cite{khurana2023point}, which proposes to use point cloud forecasting as a proxy for occupancy forecasting. It focuses on voxelized point cloud data as input, resulting in sparse and partially observed voxels as shown in \Cref{fig:related_task}. Forecasting these voxels essentially models particle motions under the Lagrangian specification. In comparison, OCF aims to forecast a sequence of dense and completed occupancy grids under the Eulerian specification, which provides a more comprehensive perception in complex environments.

\textbf{Perception dataset in autonomous driving.} Numerous public datasets for autonomous driving have been released in recent years, giving rise to research in multiple tasks~\cite{deng2021voxel,qi2017pointnet,chen2023deepmapping2}. Some datasets invest substantial resources to produce high-quality human-annotated point-level semantic labels~\cite{behley2019semantickitti,caesar2020nuscenes,sun2020scalability,liao2022kitti}, while others provide less resource-intensive labels such as bounding boxes and instance IDs~\cite{houston2021one,huang2018apolloscape,chang2019argoverse,mao2021one,geyer2020a2d2,pham20203d}. Some work delve to adapt existing public dataset to specific tasks and provide detailed benchmarks for SSC~\cite{li2023sscbench,tian2023occ3d}. In terms of forecasting, \cite{wilson2023argoverse} incorporates a motion forecasting dataset focused on object-level forecasting, and its unannotated LiDAR dataset is used in~\cite{khurana2023point} for point/occupancy forecasting. In this paper, we generate the OCFBench dataset by processing and adapting publicly available datasets for the OCF task. We deliberately select datasets with only instance labels to showcase the scalability of OCF, yet our pipeline is inherently compatible with semantic labels with zero modifications.


\section{Problem formulation}
We aim to complete and forecast the scene represented as occupancy grids within a spatial-temporal range, given point cloud inputs. Each voxel is either \textit{occupied} or \textit{unoccupied} at a given time stamp. In practice, voxels for static objects are always occupied, and free voxels are always unoccupied across the whole forecasting range. Within the short time range, voxels for dynamic objects mostly changes only once from one status to another. The status change of neighboring voxels are correlated because of the continuous movement of dynamic objects. These spatial-temporal correlation between voxels makes the problem predictable and can be learned from large scale data. We use a neural network $f_\theta$ to learn the spatial-temporal correlation from the training data and be able to generalize the forecasting. As shown in \Cref{fig:ocf}, given input as a consecutive sequence of point clouds represented as voxel grids $\mathcal{P}=\{P_{t}\}_{t=-T}^0$, the output is a sequence of completed voxel grids $\mathcal{Y}=f_\theta(\mathcal{P})=\{Y_{t}\}_{t=0}^T$.

It is worth noting that this formulation distinguishes from existing point/occupancy forecasting literature in that OCF requires forecasting the complected dense voxel grids. Statistically, the completed voxel grids have nearly 18 times more occupied voxels than the sparse ones. As shown in \Cref{fig:related_task}, this more demanding task aims to provide a comprehensive representation of the environment, while mitigating the influence of sensor intrinsics and extrinsics. This Eulerian specification of the environment underscores the need for a more efficient and robust perception.



\section{DATA CURATION}
\subsection{Data processing pipeline}
\textbf{Overview.} One of the key hurdles preventing the development of occupancy-based perception is the difficulty of capturing ground truth occupancy in the real world. Although LiDAR sensors are able to provide accurate surface points, the density-to-cost trade-off makes it impossible to obtain dense occupancy for all objects and structures in the environment. Moreover, because the sensor depends on light detection and ranging, occlusion poses an additional challenge, especially in autonomous driving scenarios where numerous dynamic objects lead to large areas of occlusions. In the following paragraphs, we introduce the challenges in processing data for OCF and the corresponding solutions in our data processing pipeline.

\begin{figure}[t]
    \centering
    \subfigure[Dynamic objects synchronization]
    {
        \includegraphics[width=\linewidth]{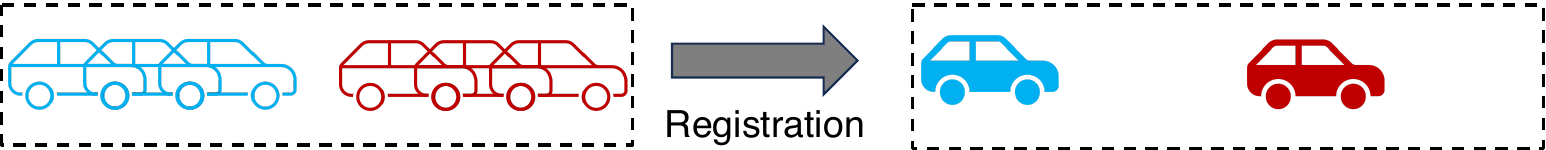}
        \label{fig:object_sync}
    }
    \subfigure[Unknown voxel exclusion]{
        \includegraphics[width=\linewidth]{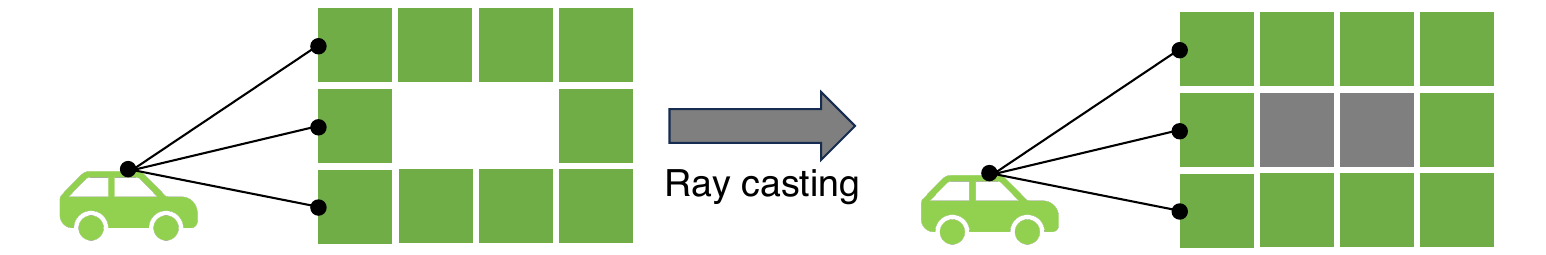}
        \label{fig:unknown_exclusion}
    }
    \subfigure[Coordinate unification]{
        \includegraphics[width=\linewidth]{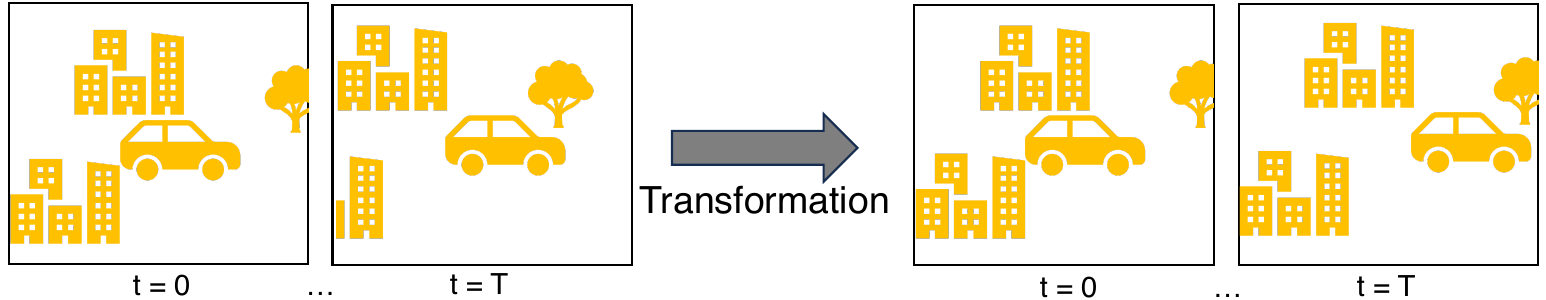}
        \label{fig:coordinate_uni}
    }
    \vspace{-4mm}
    \caption{\textbf{Steps in data processing}. (a)~Dynamic-object-synchronization addresses spatial-temporal tubes by registering each object individually. (b)~Unknown voxels are handled by running a ray-casting algorithm to find out unknown voxels and ignore them for supervision and evaluation (c)~Changing sensor extrinsics is compensated by unifying different coordinate frames to the $t=0$ frame.}
    \label{fig:dataset}
    \vspace{-4mm}
\end{figure}

\textbf{``Spatial-temporal tubes''.} 
SemanticKITTI~\cite{behley2019semantickitti} proposes to superimpose consecutive LiDAR sweeps to create a dense occupancy grid. However, this induces problems in complex environments with dynamic objects. As shown on the left part of \Cref{fig:object_sync}, it introduces the ``spatial-temporal tubes'' from moving objects. Occ3D and SSCBench~\cite{li2023sscbench,tian2023occ3d} compensate for this limitation by \textit{dynamic-object-synchronization} that employs semantic and instance labels to individually register the same objects. We follow this practice but without access to semantic labels. Unlike their utilization of the semantic labels for synchronization, we only use instance labels and bounding boxes to identify the same objects. 

\textbf{Unknown voxels.} {\Cref{fig:unknown_exclusion} and~\Cref{fig:ocf} (lower right corner of outputs) shows another problem of unknown voxels.} Some voxels that are scanned in none of the aggregated frames will be regarded as \texttt{free} while their occupancy status is actually \texttt{unknown}. {We follow previous work~\cite{tian2023occ3d,li2023sscbench} to run a ray casting algorithm to find out all unknown voxels and ignore them for supervision and evaluation.} Note that while prior-knowledge-based inference may offer extra information on the occupancy of unknown voxels, we deliberately avoid so in order to ensure the fidelity of ground truth and minimize errors and biases originating from these steps.

\textbf{Egomotion of vehicle.} One problem remains when adapting previous solutions to generate our OCFBench dataset. As shown in \Cref{fig:coordinate_uni}, consecutive LiDAR frames are scanned when the ego-vehicle is moving, resulting in changing extrinsics. This is not supposed to be modeled according to the problem formulation. We address this problem by transforming all input and ground truth frames to the same coordinate frame with the $t=0$ frame. As shown on the right part in \Cref{fig:coordinate_uni}, the static environment (such as buildings and vegetation) should not move across the forecast frames, while only the dynamic objects should move.

\begin{table*}[ht]\centering
\caption{\textbf{Results on OCFBench-Lyft}. We report the performances w.r.t. different temporal ranges. The top three performances with the same i/o setup are marked by \tbr{red}, \tbg{green}, and \tbb{blue} respectively. All metrics except mAP are with threshold $=0.5$. \\ $^*$PCF model structure without depth rendering.}
\label{tab:quant-lyft}
\vspace{-2mm}
\resizebox{0.8\linewidth}{!}{%
\begin{tabular}{l|ccc|ccc|ccc|ccc}\toprule
\textbf{Method} &\multicolumn{3}{c|}{\textbf{PCF$^*$~\cite{khurana2023point}}} &\multicolumn{3}{c|}{\textbf{ConvLSTM}} &\multicolumn{3}{c|}{\textbf{Conv3D}} &\multicolumn{3}{c}{\textbf{Conv3D+sIoU}} \\\midrule
\textbf{Input/Output} &\textbf{5/5} &\textbf{5/10} &\textbf{10/10}&\textbf{5/5} &\textbf{5/10} &\textbf{10/10}&\textbf{5/5} &\textbf{5/10} &\textbf{10/10}&\textbf{5/5} &\textbf{5/10} &\textbf{10/10}\\\midrule
\textbf{mIoU (\%)}&57.10  &56.52 & \tbb{59.09} & \tbb{59.66} & \tbb{56.56}& {59.00} & \tbg{60.05} &\tbg{59.61} & \tbg{62.07} & \tbr{61.11} & \tbr{60.59}& \tbr{62.71}\\
\textbf{mAP (\%)} &\tbb{82.65} & \tbb{82.01}& \tbb{84.14} & 81.88 & 81.49& 83.42 & \tbr{84.70} & \tbr{84.21}& \tbr{86.06} & \tbg{84.49} & \tbg{83.91}& \tbg{85.74}\\
\textbf{Precision (\%)}  &\tbg{79.83} &\tbg{79.61} & \tbr{81.26} & \tbb{78.67} &\tbb{78.70} & \tbb{79.66} & \tbr{79.88} &\tbr{79.83} & \tbg{80.87} & 76.69 &76.64 & 77.47\\
\textbf{Recall (\%)} &66.74 & 65.98& 68.37 & \tbb{67.41} &\tbb{66.78} & \tbb{69.43} & \tbg{70.73} & \tbg{70.14}& \tbg{72.68} & \tbr{75.03} & \tbr{74.26}& \tbr{76.71}\\
\textbf{F1 (\%)} &\tbb{72.57} & 72.09& \tbb{74.14} & 72.49 &\tbb{72.14} & 74.10 & \tbg{74.92} & \tbg{74.57}& \tbg{76.47} & \tbr{75.76} & \tbr{75.34}& \tbr{76.96}\\
\bottomrule
\end{tabular}
}
\end{table*}

\begin{table*}[t]\centering
\caption{\textbf{Results on OCFBench-Argoverse}. All metrics and legends follow those in \Cref{tab:quant-lyft}}
\label{tab:quant-argoverse}
\vspace{-2mm}
\resizebox{0.8\linewidth}{!}{%
\begin{tabular}{l|ccc|ccc|ccc|ccc}\toprule
\textbf{Method} &\multicolumn{3}{c|}{\textbf{PCF$^*$~\cite{khurana2023point}}} &\multicolumn{3}{c|}{\textbf{ConvLSTM}} &\multicolumn{3}{c|}{\textbf{Conv3D}} &\multicolumn{3}{c}{\textbf{Conv3D+sIoU}} \\\midrule
\textbf{Input/Output} &\textbf{5/5} &\textbf{5/10} &\textbf{10/10}&\textbf{5/5} &\textbf{5/10} &\textbf{10/10}&\textbf{5/5} &\textbf{5/10} &\textbf{10/10}&\textbf{5/5} &\textbf{5/10} &\textbf{10/10}\\\midrule
\textbf{mIoU (\%)}& \tbb{53.00} &\tbb{52.88} & {54.05} & 52.84 &52.68 & \tbb{54.33} & \tbg{54.36} & \tbg{54.42}& \tbg{55.41} & \tbr{55.41} &\tbr{54.86} & \tbr{56.70}\\
\textbf{mAP (\%)} &\tbb{78.38} & \tbb{78.35}& \tbb{79.41} & 75.45 & 75.66& {77.67} & \tbg{79.23} & \tbg{79.24}& \tbg{80.17} & \tbr{80.07} & \tbr{79.38}& \tbr{81.16}\\
\textbf{Precision (\%)}  & \tbg{73.13} &\tbg{73.74} & \tbg{73.95} & \tbb{71.95} &\tbb{72.31} & {72.99} & \tbr{73.46} &\tbr{74.01} & \tbr{74.50} & 71.13 & 71.42 &\tbb{73.09}\\
\textbf{Recall (\%)} &66.55 & 65.78& {67.32} & \tbb{67.35} &\tbb{66.72}  & \tbb{68.63}& \tbg{68.26} &\tbg{67.90} & \tbg{68.88} & \tbr{71.98} & \tbr{70.79}&\tbr{72.03} \\
\textbf{F1 (\%)} &\tbb{68.89} & \tbb{68.79}& {69.78}& 68.72 &68.59 & \tbb{69.99} & \tbg{70.02} & \tbg{70.07}& \tbg{70.87} & \tbr{70.87} &\tbr{70.40} &\tbr{71.84} \\
\bottomrule
\end{tabular}
}
\end{table*}

\subsection{OCFBench dataset}
\textbf{Overview.} To enable developing models for OCF, we establish a large-scale dataset, termed \textit{OCFBench}, for training and evaluation. We harness several existing public datasets for autonomous driving perception. We process the raw data and unify them into the same format for easier training and pave the way for research in cross-domain adaptation.

{\textbf{Dataset selection.} We select the  Lyft (Woven Planet) Level-5~\cite{houston2021one}, Argoverse~\cite{chang2019argoverse}, and ApolloScape~\cite{huang2018apolloscape} dataset as our building blocks for the OCFBench dataset. As our data processing steps only require the raw data to possess instance labels and sensor poses, we can take good advantage of the diverse data in these datasets while not hampered by the lack of point-level semantic annotations. It showcases the scalability of the OCF task because of its compatibility with numerous publicly available datasets. As our data processing pipeline is also naturally compatible with semantic labels, in our future maintenance of the dataset, we will also include the widely-used nuScenes~\cite{caesar2020nuscenes} and Waymo~\cite{sun2020scalability} dataset.}

\textbf{OCFBench-Lyft.} The Lyft Level-5 dataset~\cite{houston2021one} was originally designed for developing motion forecasting and planning solutions. It claims to provide driving data over $1,000$ hours with 162k scenes, while only 180 scenes are publicly available. It supports many tasks including semantic segmentation, trajectory prediction, and imitation learning~\cite{zhang2021end,ivanovic2022heterogeneous,zhou2022cross}. We utilize the 180 publicly available scenes and apply our processing pipeline. We allocate 120 scenes for training, 30 scenes for validation, and 30 scenes for testing respectively. In terms of the number of frames, we have 14988/3631/3750 frames for training/validation/testing. The final dataset is composed of $22,369$ frames ($\sim$22k) in total. 

\textbf{OCFBench-Argoverse.} The Argoverse dataset~\cite{chang2019argoverse} is dedicated for 3D tracking and forecasting. It features its mined trajectory for motion forecasting. It inspires multiple research on trajectory prediction~\cite{gu2021densetnt,salzmann2020trajectron++,ettinger2021large}. While a motion forecasting dataset is provided in Argoverse, it is based on HD maps and incompatible with our formulation for OCF. Instead, we utilize the 3D tracking dataset that is composed of 89 publicly available segments and divide them into 50/15/24 segments for training/validation/testing respectively. This distribution results in a total of $13,099$ frames ($\sim$13k), with $7,005$ frames allocated for training, $2,180$ frames for validation, and $3,914$ frames for testing.

\textbf{OCFBench-ApolloScape.} The ApolloScape~\cite{huang2018apolloscape} dataset supports various tasks for autonomous driving research, including localization, semantic parsing, and instance segmentation. Many algorithms conduct experiments on the dataset for depth estimation or semantic segmentation~\cite{ranftl2021vision,sakaridis2021acdc}. We process the 52 scenes of its tracking data and generate the OCFBench-ApolloScape dataset that includes $4,389$ frames, where the train/valid/test scene split is set to 40/6/6, resulting in 3,469 frames for training, 583 frames for validation, and 337 frames for testing.

\begin{table*}[t]\centering
\caption{\textbf{Results on OCFBench-ApolloScape}. All metrics and legends follow those in \Cref{tab:quant-lyft}}
\label{tab:quant-apolloscape}
\vspace{-2mm}
\resizebox{0.8\linewidth}{!}{%
\begin{tabular}{l|ccc|ccc|ccc|ccc}\toprule
\textbf{Method} &\multicolumn{3}{c|}{\textbf{PCF$^*$~\cite{khurana2023point}}} &\multicolumn{3}{c|}{\textbf{ConvLSTM}} &\multicolumn{3}{c|}{\textbf{Conv3D}} &\multicolumn{3}{c}{\textbf{Conv3D+sIoU}} \\\midrule
\textbf{Input/Output} &\textbf{5/5} &\textbf{5/10} &\textbf{10/10}&\textbf{5/5} &\textbf{5/10} &\textbf{10/10}&\textbf{5/5} &\textbf{5/10} &\textbf{10/10}&\textbf{5/5} &\textbf{5/10} &\textbf{10/10}\\\midrule
\textbf{mIoU (\%)}& 58.70 & \tbb{58.77} & \tbb{59.61} & \tbb{58.73} & 58.24 & 59.23 & \tbg{61.16} & \tbg{61.13} & \tbg{61.30} & \tbr{61.58} & \tbr{61.66} & \tbr{62.42}\\
\textbf{mAP (\%)} & \tbb{83.39} & \tbb{83.30} & \tbb{84.02} & 82.34 & 81.93 & 83.03 & \tbg{84.84} & \tbr{84.80} & \tbg{85.12} & \tbr{84.92} & \tbg{84.47} & \tbr{85.30}\\
\textbf{Precision (\%)} & \tbr{80.94} & \tbr{81.33} & \tbr{81.46} & \tbb{79.02} & \tbb{78.95} & \tbb{79.69} & \tbg{80.70} & \tbg{80.82} & \tbg{81.39} & 77.20 & 76.93 & 78.06\\
\textbf{Recall (\%)} & 67.99 & 67.88 & 68.85 & \tbb{69.50} & \tbb{68.81} & \tbb{69.61} & \tbg{71.40} & \tbg{71.26} & \tbg{71.05} & \tbr{74.98} & \tbr{75.31} & \tbr{75.43}\\
\textbf{F1 (\%)} & 73.60 & \tbb{73.63} & \tbb{74.27} & \tbb{73.71} & 73.28 & 74.05 & \tbg{75.55} & \tbg{75.54} & \tbg{75.53} & \tbr{75.92} & \tbr{75.97} & \tbr{76.52}\\
\bottomrule
\end{tabular}
}
\end{table*}

\section{EXPERIMENT}
\subsection{Benchmark methods}
\textbf{PCF.} The model structure is derived from \cite{khurana2023point}, which is in turn adapted from~\cite{khurana2022differentiable} and~\cite{zeng2019end}. In our implementation, we omit the depth rendering module to make it compatible with the OCF problem formulation. The model has a simple convolution-based encoder-decoder structure. One technique used in~\cite{khurana2023point} is to reshape the tensor and concatenate the temporal dimension onto the height, thus fitting the 4D voxel tensor into 2D convolutional layers. Note that our adaptation is specifically trained under the OCF problem formulation, utilizing previously mentioned loss functions. As such, it is not directly comparable to the original model in \cite{khurana2023point}.

\textbf{Improvements.} Building upon OCF, we explore enhancements through two additional models: ConvLSTM~\cite{shi2015convolutional} and Conv3D~\cite{tran2015learning}. In our implementation of ConvLSTM, we employ the convolutional blocks in the OCF but remove the concatenation step. We use a shared 2D convolution encoder for all input frames, and recurrently feed temporal features to the LSTM module. For Conv3D, we implement by replacing the 2D convolutional layers in the base structure with 3D convolutional layers. While this indicates a larger memory footprint, our experiment shows that the training process is able to fit in one single GPU.

In addition to model structures, we also explore the impact of the soft-IoU loss function. Soft-IoU is introduced in~\cite{huang2019batching} primarily as a metric to better evaluate the confidence of model predictions. As a side effect, the softness makes the metric differentiable and can be utilized as a loss function. The proposed loss function is:
\begin{equation}
    \mathcal{L}(y,\Tilde{y})=-\frac{1}{|C|}\sum_C \frac{\sum_V y\cdot\Tilde{y}}{\sum_V y+\Tilde{y}-y\cdot\Tilde{y}},
\end{equation}
where $C$ is the mini-batch, $V$ is the set of voxels in one sample, $y$ is the ground truth occupancy represented as $\{0,1\}$, and $\Tilde{y}$ is the predicted occupancy probability for each voxel. Note that this loss function not only incorporates the idea of IoU, but also enables the model to have a more confident prediction. In our experiment, we train the 3D convolution model using the sum of BCE and soft-IoU loss.

\subsection{Evaluation metric}
\textbf{mIoU.} We adopt the intersection over union (IoU) to measure the OCF performance following the convention in SSC literature. Unlike computing the mean of per-class IoU, we compute the mean of per-frame geometric IoU\footnote{Technically, it should be referred to as Jaccard index in the binary classification case but we keep the term consistent with SSC literature.} to assess the overall OCF performance along the temporal dimension.

\textbf{mAP.} Because the model predicts the continuous probability for each voxel, a binary evaluation metric is unable to evaluate the performance with different thresholds. We use mean average precision (mAP) as one of the metrics to present an overall and comprehensive evaluation of the confidence of the model prediction. It is calculated by averaging the area under the precision-recall curve for each frame. It provides a comprehensive measure of a model's performance by considering both precision and recall.

\textbf{Precision, recall, and F1 score.} These are widely used metrics for binary classification. We include these metrics for a more diverse evaluation of each method.

\begin{figure*}[ht]
    \centering
    \includegraphics[width=0.95\linewidth]{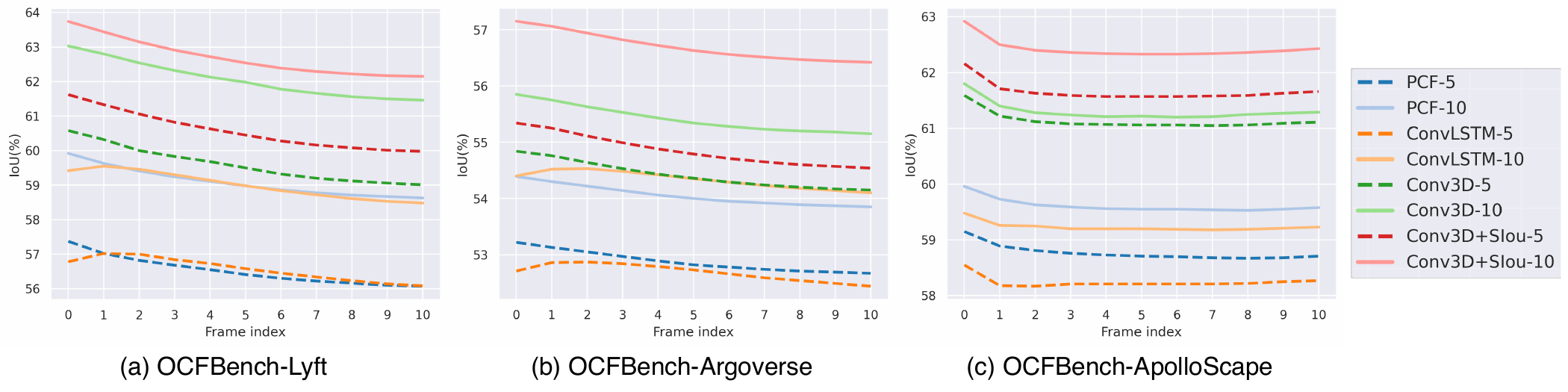}
    \vspace{-2mm}
    \caption{\textbf{Performance degradation w.r.t. time}. We show the per-frame IoU for each method when forecasting 10 future frames with 5/10 input frames.}
    \label{fig:exp_temporal}
    \vspace{-2mm}
\end{figure*}

For all metrics above, we chose three setups with different input/output \textbf{temporal ranges}, namely 5/5, 5/10, and 10/10 for input/output frames. This design provides insightful analysis of performance relative to the temporal range, which is crucial for many downstream tasks including planning and navigation~\cite{hu2023planning}. Furthermore, by comparing the performance between 5/5 and 5/10, we should be able to see the challenge of forecasting longer sequences with the same input. On the other hand, comparing 5/10 with 10/10, we can witness the benefits of more input frames.

\subsection{Common results on all datasets}

\textbf{Different model architecture.} \Cref{tab:quant-lyft,tab:quant-argoverse,tab:quant-apolloscape} show the performance of different methods on our three datasets. We can observe that the Conv3D architecture generally outperforms ConvLSTM and PCF on all temporal ranges and evaluation metrics, especially in terms of mIoU and mAP, two of the most comprehensive ones. It shows the advantage of 3D convolutional layer in processing 3D data because it is capable of modeling the spatial contexts in 3D structure. This improvement is expected, considering that PCF directly concatenates temporal dimension onto the height dimension to fit 4D data into a 2D convolutional layer. On the other hand, while we don't observe significant improvements from ConvLSTM, it features its constant number of parameters regardless of the length of the input and output. In general, we can summarize that there is a trade-off between model performance and memory requirement in this thread of model architectures. Detailed discussion is provided in \cref{sec:specs}.

\textbf{Soft-IoU loss.} The Conv3D model trained with additional soft-IoU loss achieves better evaluation results on almost all setups. The improvement is most significant on the mIoU metric, mainly because soft-IoU loss enhances the prediction confidence of the model and encourages. While there is a slight negative impact on precision, we believe this is a less concerning result from safety perspective, as false positive prediction doesn't deteriorate safety-critical downstream tasks.

\textbf{Various output intervals with same input.}
All methods experience a decrease in performance when forecasting longer sequences with the same input due to increased complexity and uncertainty in longer temporal ranges. Nevertheless, note that the degradation from 5/5 to 5/10 is not significant (less than $1\%$ for all methods across all datasets). This can be explained by the disparity between static and dynamic voxels. Given the relatively small proportion of dynamic objects within the scene, the adverse impact of mis-predictions on the overall IoU can be minimal.

\begin{figure}[t]
    \centering
    \includegraphics[width=\linewidth]{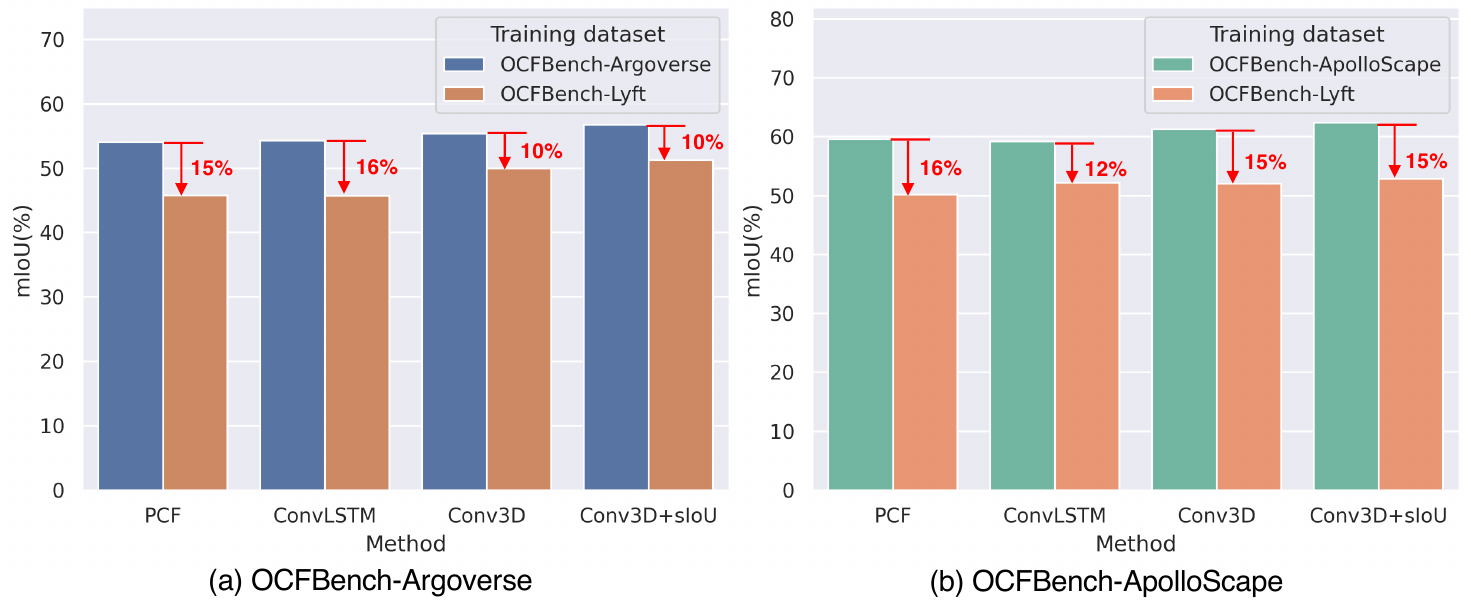}
    \caption{\textbf{Results for cross-domain evaluation}. All methods were trained on OCFBench-Lyft and tested on the other two datasets with the 10/10 input/output setup. The degradation in percentage is marked on the chart.}
    \label{fig:cross_adaptation}
    \vspace{-2mm}
\end{figure}

\textbf{Various input intervals with same output.}
Comparing the performance between 5/10 and 10/10, we can observe that both mIoU and mAP increase when more input frames were provided. This could be explained by the following reasons: 1) With more input frames, the model has access to a larger temporal context, allowing it to capture richer information and make more accurate predictions. 2) Increasing the number of input frames help mitigate the impact of noise in individual frames, which is typical for LiDAR sensors. With more input frames, the model can rely on the consistency among multiple frames to make reliable predictions.

\textbf{Temporal degradation.}
As shown in~\Cref{fig:exp_temporal}, as we extend the forecasting horizon further into the future, all the methods demonstrate a noticeable decline in per-frame performance. This can be attributed to the inherent challenge of accurately capturing long-term trends and effectively accounting for unforeseen events that may manifest in the distant future. Notably, we can observe that the performance gains achieved through advancements in model architecture surpass those obtained by simply extending the input sequence length. This observation underscores the significance of ongoing research and development efforts aimed at refining model architectures to address this challenging problem.

\textbf{Cross-domain consistency.}
In synthesizing the findings from the results on the three different datasets, it is noteworthy that the performance trends are largely invariant across these diverse settings. This consistency suggests the OCF problem is not extremely sensitive to different specifications, considering all three of our datasets have different sensor setups. While subtle variations do manifest, they do not present a significant deviation that would warrant dataset-specific adaptations. This observation is impactful for real-world applications, where model versatility across different sensor data and environmental contexts is crucial. Unlike the SSC that is sensitive to sensor setups~\cite{li2023sscbench}, this observation provides a strong foundation for future work to enhance the generalizability of models.

\subsection{Results for cross-domain adaptation}
To better examine the cross-domain adaptation of the baseline methods, we conduct experiments for cross-validation. Because the OCFBench-Lyft dataset has the most number of sequences and frames compared to the other two datasets, we use the model trained on OCFBench-Lyft and test it on both OCFBench-Argoverse and OCFBench-ApolloScape. As depicted in~\Cref{fig:cross_adaptation}, all methods exhibit a deterioration in performance (more than $10\%$). Specifically, the performance dropped notably (PCF mIoU: $54.05\rightarrow45.75$) when tested on OCFBench-Argoverse. This is similar for OCFBench-ApolloScape (PCF mIoU: $59.61\rightarrow50.19$). 

Note that this decline in performance is observed in spite of the fact that OCFBench-Lyft has approximately 2 times more training data than OCFBench-Argoverse and 4 times more than OCFBench-ApolloScape. This trend underscores one of the motivations to propose the task of OCF: to utilize more publicly available datasets to achieve environment perception that is robust to domain gaps.

\begin{table}[t]
    \centering
    \caption{\textbf{Model specs}. All reported specs are recorded when training on an A100 GPU with the OCFBench-Lyft dataset and a batch size of 2. Tr. stands for training and Inf. for inference.}
    \vspace{-2mm}
    \label{tab:model_specs}
    \resizebox{\linewidth}{!}{%
    \begin{tabular}{l|ccc|ccc|ccc}
        \toprule
        \textbf{Method} &\multicolumn{3}{c|}{\textbf{PCF$^*$~\cite{khurana2023point}}} &\multicolumn{3}{c|}{\textbf{ConvLSTM}} &\multicolumn{3}{c}{\textbf{Conv3D}} \\ \midrule
        \textbf{Input/ourput} & \textbf{5/5} & \textbf{5/10} & \textbf{10/10} & \textbf{5/5} & \textbf{5/10} & \textbf{10/10} & \textbf{5/5} & \textbf{5/10} & \textbf{10/10} \\ \midrule
        \textbf{Tr. time (s/batch)} & 0.61 & 0.89 & 1.07 & 0.84 & 1.26 & 1.61 & 1.53 & 1.74 & 1.78 \\ 
        \textbf{Inf. time (s/batch)} & 0.60 & 0.84 & 1.05 & 0.70 & 0.48 & 1.58 & 0.91 & 1.23 & 1.73 \\ 
        \textbf{MACs (T)} & 0.27 & 0.37 & 0.51 & 1.74 & 2.44 & 3.24 & 6.39 & 6.47 & 6.51 \\ 
        \textbf{Parameter (M)} & 8.2 & 8.5 & 9.0 & 12 & 12 & 12 & 34 & 34 & 34 \\ 
        \textbf{Memory (GB)} & 12 & 15 & 17 & 14 & 15 & 21 & 23 & 23 & 23 \\ 
        \bottomrule
    \end{tabular}
    \vspace{-2mm}
    }
\end{table}

\subsection{Model specs.}\label{sec:specs}
\Cref{tab:model_specs} lists the model specifications and computation resources required during training. It is evident that performance gains come at the expense of increased computational complexity. The Conv3D architecture, despite outperforming the rest architectures in mIoU and mAP, imposes a substantial computational burden. It exhibits higher inference time, multiply-accumulate operations (MACs), and memory consumption, as compared to PCF and ConvLSTM. ConvLSTM, despite its moderate performance, offers a more balanced profile in terms of computational requirements. Note that a large portion of parameters ($\sim$ 9M) in ConvLSTM comes from the same encoder architecture as PCF. Despite its moderate performance, ConvLSTM has the advantage of a constant number of parameters that is independent of input/output length. These disparities implicate the potential to optimize the trade-off between computational resources and performance efficacy. This is particularly essential for real-time and embodied applications in which resource limitations can be a significant constraint.

\section{Conclusion}
\textbf{Summary.} In this paper, we propose OCF, a new spatial-temporal perception task in the context of autonomous driving. It poses challenges requiring joint reconstruction, hallucination, and prediction. We curate a large-scale dataset termed OCFBench to facilitate the training and evaluation. Our experiments on existing show the challenge and potential of solving the problem. We hope our work will inspire future research in this field.

\textbf{Limitations and future work.} The problem formulation of OCF and our experiments only considers point cloud sequences as input. We will include camera-based methods in future work. Our data curation relies on sensor poses and bonding box labels for registration. Inaccurate poses and annotations can errors in ground truth when extending to higher-resolution data. Our future development of the dataset will include nuScenes~\cite{caesar2020nuscenes} and Waymo~\cite{sun2020scalability}.

\bibliographystyle{ieeetran}
\bibliography{IEEEabrv,IEEEexample}

\begin{thebibliography}{10}
\providecommand{\url}[1]{#1}
\csname url@rmstyle\endcsname
\providecommand{\newblock}{\relax}
\providecommand{\bibinfo}[2]{#2}
\providecommand\BIBentrySTDinterwordspacing{\spaceskip=0pt\relax}
\providecommand\BIBentryALTinterwordstretchfactor{4}
\providecommand\BIBentryALTinterwordspacing{\spaceskip=\fontdimen2\font plus
\BIBentryALTinterwordstretchfactor\fontdimen3\font minus
  \fontdimen4\font\relax}
\providecommand\BIBforeignlanguage[2]{{%
\expandafter\ifx\csname l@#1\endcsname\relax
\typeout{** WARNING: IEEEtran.bst: No hyphenation pattern has been}%
\typeout{** loaded for the language `#1'. Using the pattern for}%
\typeout{** the default language instead.}%
\else
\language=\csname l@#1\endcsname
\fi
#2}}

\bibitem{shi2023grid}
Y.~Shi, K.~Jiang, J.~Li, J.~Wen, Z.~Qian, M.~Yang, K.~Wang, and D.~Yang,
  ``Grid-centric traffic scenario perception for autonomous driving: A
  comprehensive review,'' \emph{arXiv preprint arXiv:2303.01212}, 2023.

\bibitem{li2022bevformer}
Z.~Li, W.~Wang, H.~Li, E.~Xie, C.~Sima, T.~Lu, Y.~Qiao, and J.~Dai,
  ``Bevformer: Learning bird’s-eye-view representation from multi-camera
  images via spatiotemporal transformers,'' in \emph{ECCV}.\hskip 1em plus
  0.5em minus 0.4em\relax Springer, 2022, pp. 1--18.

\bibitem{hu2021fiery}
A.~Hu, Z.~Murez, N.~Mohan, S.~Dudas, J.~Hawke, V.~Badrinarayanan, R.~Cipolla,
  and A.~Kendall, ``Fiery: Future instance prediction in bird's-eye view from
  surround monocular cameras,'' in \emph{CVPR}, 2021, pp. 15\,273--15\,282.

\bibitem{zhang2022beverse}
Y.~Zhang, Z.~Zhu, W.~Zheng, J.~Huang, G.~Huang, J.~Zhou, and J.~Lu, ``Beverse:
  Unified perception and prediction in birds-eye-view for vision-centric
  autonomous driving,'' \emph{arXiv preprint arXiv:2205.09743}, 2022.

\bibitem{hu2023planning}
Y.~Hu, J.~Yang, L.~Chen, K.~Li, C.~Sima, X.~Zhu, S.~Chai, S.~Du, T.~Lin,
  W.~Wang, \emph{et~al.}, ``Planning-oriented autonomous driving,'' in
  \emph{CVPR}, 2023, pp. 17\,853--17\,862.

\bibitem{song2017semantic}
S.~Song, F.~Yu, A.~Zeng, A.~X. Chang, M.~Savva, and T.~Funkhouser, ``Semantic
  scene completion from a single depth image,'' in \emph{CVPR}, 2017, pp.
  1746--1754.

\bibitem{roldao2020lmscnet}
L.~Roldao, R.~de~Charette, and A.~Verroust-Blondet, ``Lmscnet: Lightweight
  multiscale 3d semantic completion,'' in \emph{3DV}.\hskip 1em plus 0.5em
  minus 0.4em\relax IEEE, 2020, pp. 111--119.

\bibitem{cao2022monoscene}
A.-Q. Cao and R.~de~Charette, ``Monoscene: Monocular 3d semantic scene
  completion,'' in \emph{CVPR}, 2022, pp. 3991--4001.

\bibitem{li2023voxformer}
Y.~Li, Z.~Yu, C.~Choy, C.~Xiao, J.~M. Alvarez, S.~Fidler, C.~Feng, and
  A.~Anandkumar, ``Voxformer: Sparse voxel transformer for camera-based 3d
  semantic scene completion,'' in \emph{CVPR}, 2023, pp. 9087--9098.

\bibitem{khurana2022differentiable}
T.~Khurana, P.~Hu, A.~Dave, J.~Ziglar, D.~Held, and D.~Ramanan,
  ``Differentiable raycasting for self-supervised occupancy forecasting,'' in
  \emph{ECCV}.\hskip 1em plus 0.5em minus 0.4em\relax Springer, 2022, pp.
  353--369.

\bibitem{khurana2023point}
T.~Khurana, P.~Hu, D.~Held, and D.~Ramanan, ``Point cloud forecasting as a
  proxy for 4d occupancy forecasting,'' in \emph{CVPR}, 2023, pp. 1116--1124.

\bibitem{fei2022comprehensive}
B.~Fei, W.~Yang, W.-M. Chen, Z.~Li, Y.~Li, T.~Ma, X.~Hu, and L.~Ma,
  ``Comprehensive review of deep learning-based 3d point cloud completion
  processing and analysis,'' \emph{{IEEE} Trans. Intell. Transport. Syst.},
  2022.

\bibitem{wang2020point}
X.~Wang, M.~H. Ang, and G.~H. Lee, ``Point cloud completion by learning shape
  priors,'' in \emph{IROS}.\hskip 1em plus 0.5em minus 0.4em\relax IEEE, 2020,
  pp. 10\,719--10\,726.

\bibitem{Reza2022occupancy}
R.~Mahjourian, J.~Kim, Y.~Chai, M.~Tan, B.~Sapp, and D.~Anguelov, ``Occupancy
  flow fields for motion forecasting in autonomous driving,'' \emph{IEEE Robot.
  Automat. Lett.}, vol.~7, no.~2, pp. 5639--5646, 2022.

\bibitem{behley2019semantickitti}
J.~Behley, M.~Garbade, A.~Milioto, J.~Quenzel, S.~Behnke, C.~Stachniss, and
  J.~Gall, ``Semantickitti: A dataset for semantic scene understanding of lidar
  sequences,'' in \emph{CVPR}, 2019, pp. 9297--9307.

\bibitem{tian2023occ3d}
X.~Tian, T.~Jiang, L.~Yun, Y.~Wang, Y.~Wang, and H.~Zhao, ``Occ3d: A
  large-scale 3d occupancy prediction benchmark for autonomous driving,''
  \emph{arXiv preprint arXiv:2304.14365}, 2023.

\bibitem{li2023sscbench}
Y.~Li, S.~Li, X.~Liu, M.~Gong, K.~Li, N.~Chen, Z.~Wang, Z.~Li, T.~Jiang, F.~Yu,
  Y.~Wang, H.~Zhao, Z.~Yu, and C.~Feng, ``Sscbench: A large-scale 3d semantic
  scene completion benchmark for autonomous driving,'' \emph{arXiv preprint
  arXiv:2306.09001}, 2023.

\bibitem{zheng2013beyond}
B.~Zheng, Y.~Zhao, J.~C. Yu, K.~Ikeuchi, and S.-C. Zhu, ``Beyond point clouds:
  Scene understanding by reasoning geometry and physics,'' in \emph{CVPR},
  2013, pp. 3127--3134.

\bibitem{ramakrishnan2018sidekick}
S.~K. Ramakrishnan and K.~Grauman, ``Sidekick policy learning for active visual
  exploration,'' in \emph{ECCV}, 2018, pp. 413--430.

\bibitem{song2018im2pano3d}
S.~Song, A.~Zeng, A.~X. Chang, M.~Savva, S.~Savarese, and T.~Funkhouser,
  ``Im2pano3d: Extrapolating 360 structure and semantics beyond the field of
  view,'' in \emph{CVPR}, 2018, pp. 3847--3856.

\bibitem{yang2019extreme}
Z.~Yang, J.~Z. Pan, L.~Luo, X.~Zhou, K.~Grauman, and Q.~Huang, ``Extreme
  relative pose estimation for rgb-d scans via scene completion,'' in
  \emph{CVPR}, 2019, pp. 4531--4540.

\bibitem{popovic2021volumetric}
M.~Popovi{\'c}, F.~Thomas, S.~Papatheodorou, N.~Funk, T.~Vidal-Calleja, and
  S.~Leutenegger, ``Volumetric occupancy mapping with probabilistic depth
  completion for robotic navigation,'' \emph{IEEE Robot. Automat. Lett.},
  vol.~6, no.~3, pp. 5072--5079, 2021.

\bibitem{huang2023tri}
Y.~Huang, W.~Zheng, Y.~Zhang, J.~Zhou, and J.~Lu, ``Tri-perspective view for
  vision-based 3d semantic occupancy prediction,'' in \emph{CVPR}, 2023, pp.
  9223--9232.

\bibitem{zhang2023occformer}
Y.~Zhang, Z.~Zhu, and D.~Du, ``Occformer: Dual-path transformer for
  vision-based 3d semantic occupancy prediction,'' \emph{arXiv preprint
  arXiv:2304.05316}, 2023.

\bibitem{wei2023surroundocc}
Y.~Wei, L.~Zhao, W.~Zheng, Z.~Zhu, J.~Zhou, and J.~Lu, ``Surroundocc:
  Multi-camera 3d occupancy prediction for autonomous driving,'' in
  \emph{Proceedings of the IEEE/CVF International Conference on Computer
  Vision}, 2023, pp. 21\,729--21\,740.

\bibitem{cheng2021s3cnet}
R.~Cheng, C.~Agia, Y.~Ren, X.~Li, and L.~Bingbing, ``S3cnet: A sparse semantic
  scene completion network for lidar point clouds,'' in \emph{CoRL}.\hskip 1em
  plus 0.5em minus 0.4em\relax PMLR, 2021, pp. 2148--2161.

\bibitem{xia2023scpnet}
Z.~Xia, Y.~Liu, X.~Li, X.~Zhu, Y.~Ma, Y.~Li, Y.~Hou, and Y.~Qiao, ``Scpnet:
  Semantic scene completion on point cloud,'' in \emph{CVPR}, 2023, pp.
  17\,642--17\,651.

\bibitem{sun2020novel}
X.~Sun, S.~Wang, M.~Wang, Z.~Wang, and M.~Liu, ``A novel coding architecture
  for lidar point cloud sequence,'' \emph{IEEE Robot. Automat. Lett.}, vol.~5,
  no.~4, pp. 5637--5644, 2020.

\bibitem{deng2020temporal}
D.~Deng and A.~Zakhor, ``Temporal lidar frame prediction for autonomous
  driving,'' in \emph{3DV}.\hskip 1em plus 0.5em minus 0.4em\relax IEEE, 2020,
  pp. 829--837.

\bibitem{zhang2021cloudlstm}
C.~Zhang, M.~Fiore, I.~Murray, and P.~Patras, ``Cloudlstm: A recurrent neural
  model for spatiotemporal point-cloud stream forecasting,'' in \emph{AAAI},
  vol.~35, no.~12, 2021, pp. 10\,851--10\,858.

\bibitem{weng2022s2net}
X.~Weng, J.~Nan, K.-H. Lee, R.~McAllister, A.~Gaidon, N.~Rhinehart, and K.~M.
  Kitani, ``S2net: Stochastic sequential pointcloud forecasting,'' in
  \emph{ECCV}.\hskip 1em plus 0.5em minus 0.4em\relax Springer, 2022, pp.
  549--564.

\bibitem{ye2021tpcn}
M.~Ye, T.~Cao, and Q.~Chen, ``Tpcn: Temporal point cloud networks for motion
  forecasting,'' in \emph{CVPR}, 2021, pp. 11\,318--11\,327.

\bibitem{mersch2022self}
B.~Mersch, X.~Chen, J.~Behley, and C.~Stachniss, ``Self-supervised point cloud
  prediction using 3d spatio-temporal convolutional networks,'' in
  \emph{CoRL}.\hskip 1em plus 0.5em minus 0.4em\relax PMLR, 2022, pp.
  1444--1454.

\bibitem{weng2021inverting}
X.~Weng, J.~Wang, S.~Levine, K.~Kitani, and N.~Rhinehart, ``Inverting the pose
  forecasting pipeline with spf2: Sequential pointcloud forecasting for
  sequential pose forecasting,'' in \emph{CoRL}.\hskip 1em plus 0.5em minus
  0.4em\relax PMLR, 2021, pp. 11--20.

\bibitem{schreiber2019long}
M.~Schreiber, S.~Hoermann, and K.~Dietmayer, ``Long-term occupancy grid
  prediction using recurrent neural networks,'' in \emph{ICRA}.\hskip 1em plus
  0.5em minus 0.4em\relax IEEE, 2019, pp. 9299--9305.

\bibitem{casas2021mp3}
S.~Casas, A.~Sadat, and R.~Urtasun, ``Mp3: A unified model to map, perceive,
  predict and plan,'' in \emph{CVPR}, 2021, pp. 14\,403--14\,412.

\bibitem{sadat2020perceive}
A.~Sadat, S.~Casas, M.~Ren, X.~Wu, P.~Dhawan, and R.~Urtasun, ``Perceive,
  predict, and plan: Safe motion planning through interpretable semantic
  representations,'' in \emph{ECCV}.\hskip 1em plus 0.5em minus 0.4em\relax
  Springer, 2020, pp. 414--430.

\bibitem{deng2021voxel}
J.~Deng, S.~Shi, P.~Li, W.~Zhou, Y.~Zhang, and H.~Li, ``Voxel r-cnn: Towards
  high performance voxel-based 3d object detection,'' in \emph{AAAI}, vol.~35,
  no.~2, 2021, pp. 1201--1209.

\bibitem{qi2017pointnet}
C.~R. Qi, H.~Su, K.~Mo, and L.~J. Guibas, ``Pointnet: Deep learning on point
  sets for 3d classification and segmentation,'' in \emph{CVPR}, 2017, pp.
  652--660.

\bibitem{chen2023deepmapping2}
C.~Chen, X.~Liu, Y.~Li, L.~Ding, and C.~Feng, ``Deepmapping2: Self-supervised
  large-scale lidar map optimization,'' in \emph{CVPR}, 2023, pp. 9306--9316.

\bibitem{caesar2020nuscenes}
H.~Caesar, V.~Bankiti, A.~H. Lang, S.~Vora, V.~E. Liong, Q.~Xu, A.~Krishnan,
  Y.~Pan, G.~Baldan, and O.~Beijbom, ``nuscenes: A multimodal dataset for
  autonomous driving,'' in \emph{CVPR}, 2020, pp. 11\,621--11\,631.

\bibitem{sun2020scalability}
P.~Sun, H.~Kretzschmar, X.~Dotiwalla, A.~Chouard, V.~Patnaik, P.~Tsui, J.~Guo,
  Y.~Zhou, Y.~Chai, B.~Caine, \emph{et~al.}, ``Scalability in perception for
  autonomous driving: Waymo open dataset,'' in \emph{CVPR}, 2020, pp.
  2446--2454.

\bibitem{liao2022kitti}
Y.~Liao, J.~Xie, and A.~Geiger, ``Kitti-360: A novel dataset and benchmarks for
  urban scene understanding in 2d and 3d,'' \emph{{IEEE} Trans. Pattern Anal.
  Machine Intell.}, vol.~45, no.~3, pp. 3292--3310, 2022.

\bibitem{houston2021one}
J.~Houston, G.~Zuidhof, L.~Bergamini, Y.~Ye, L.~Chen, A.~Jain, S.~Omari,
  V.~Iglovikov, and P.~Ondruska, ``One thousand and one hours: Self-driving
  motion prediction dataset,'' in \emph{CoRL}.\hskip 1em plus 0.5em minus
  0.4em\relax PMLR, 2021, pp. 409--418.

\bibitem{huang2018apolloscape}
X.~Huang, X.~Cheng, Q.~Geng, B.~Cao, D.~Zhou, P.~Wang, Y.~Lin, and R.~Yang,
  ``The apolloscape dataset for autonomous driving,'' in \emph{CVPR workshops},
  2018, pp. 954--960.

\bibitem{chang2019argoverse}
M.-F. Chang, J.~Lambert, P.~Sangkloy, J.~Singh, S.~Bak, A.~Hartnett, D.~Wang,
  P.~Carr, S.~Lucey, D.~Ramanan, \emph{et~al.}, ``Argoverse: 3d tracking and
  forecasting with rich maps,'' in \emph{CVPR}, 2019, pp. 8748--8757.

\bibitem{mao2021one}
J.~Mao, M.~Niu, C.~Jiang, H.~Liang, J.~Chen, X.~Liang, Y.~Li, C.~Ye, W.~Zhang,
  Z.~Li, \emph{et~al.}, ``One million scenes for autonomous driving: Once
  dataset,'' \emph{arXiv preprint arXiv:2106.11037}, 2021.

\bibitem{geyer2020a2d2}
J.~Geyer, Y.~Kassahun, M.~Mahmudi, X.~Ricou, R.~Durgesh, A.~S. Chung,
  L.~Hauswald, V.~H. Pham, M.~M{\"u}hlegg, S.~Dorn, \emph{et~al.}, ``A2d2: Audi
  autonomous driving dataset,'' \emph{arXiv preprint arXiv:2004.06320}, 2020.

\bibitem{pham20203d}
Q.-H. Pham, P.~Sevestre, R.~S. Pahwa, H.~Zhan, C.~H. Pang, Y.~Chen, A.~Mustafa,
  V.~Chandrasekhar, and J.~Lin, ``A 3d dataset: Towards autonomous driving in
  challenging environments,'' in \emph{ICRA}.\hskip 1em plus 0.5em minus
  0.4em\relax IEEE, 2020, pp. 2267--2273.

\bibitem{wilson2023argoverse}
B.~Wilson, W.~Qi, T.~Agarwal, J.~Lambert, J.~Singh, S.~Khandelwal, B.~Pan,
  R.~Kumar, A.~Hartnett, J.~K. Pontes, \emph{et~al.}, ``Argoverse 2: Next
  generation datasets for self-driving perception and forecasting,''
  \emph{arXiv preprint arXiv:2301.00493}, 2023.

\bibitem{zhang2021end}
Z.~Zhang, A.~Liniger, D.~Dai, F.~Yu, and L.~Van~Gool, ``End-to-end urban
  driving by imitating a reinforcement learning coach,'' in \emph{CVPR}, 2021,
  pp. 15\,222--15\,232.

\bibitem{ivanovic2022heterogeneous}
B.~Ivanovic, K.-H. Lee, P.~Tokmakov, B.~Wulfe, R.~Mcllister, A.~Gaidon, and
  M.~Pavone, ``Heterogeneous-agent trajectory forecasting incorporating class
  uncertainty,'' in \emph{IROS}.\hskip 1em plus 0.5em minus 0.4em\relax IEEE,
  2022, pp. 12\,196--12\,203.

\bibitem{zhou2022cross}
B.~Zhou and P.~Kr{\"a}henb{\"u}hl, ``Cross-view transformers for real-time
  map-view semantic segmentation,'' in \emph{CVPR}, 2022, pp. 13\,760--13\,769.

\bibitem{gu2021densetnt}
J.~Gu, C.~Sun, and H.~Zhao, ``Densetnt: End-to-end trajectory prediction from
  dense goal sets,'' in \emph{CVPR}, 2021, pp. 15\,303--15\,312.

\bibitem{salzmann2020trajectron++}
T.~Salzmann, B.~Ivanovic, P.~Chakravarty, and M.~Pavone, ``Trajectron++:
  Dynamically-feasible trajectory forecasting with heterogeneous data,'' in
  \emph{ECCV}.\hskip 1em plus 0.5em minus 0.4em\relax Springer, 2020, pp.
  683--700.

\bibitem{ettinger2021large}
S.~Ettinger, S.~Cheng, B.~Caine, C.~Liu, H.~Zhao, S.~Pradhan, Y.~Chai, B.~Sapp,
  C.~R. Qi, Y.~Zhou, \emph{et~al.}, ``Large scale interactive motion
  forecasting for autonomous driving: The waymo open motion dataset,'' in
  \emph{CVPR}, 2021, pp. 9710--9719.

\bibitem{ranftl2021vision}
R.~Ranftl, A.~Bochkovskiy, and V.~Koltun, ``Vision transformers for dense
  prediction,'' in \emph{CVPR}, 2021, pp. 12\,179--12\,188.

\bibitem{sakaridis2021acdc}
C.~Sakaridis, D.~Dai, and L.~Van~Gool, ``Acdc: The adverse conditions dataset
  with correspondences for semantic driving scene understanding,'' in
  \emph{CVPR}, 2021, pp. 10\,765--10\,775.

\bibitem{zeng2019end}
W.~Zeng, W.~Luo, S.~Suo, A.~Sadat, B.~Yang, S.~Casas, and R.~Urtasun,
  ``End-to-end interpretable neural motion planner,'' in \emph{CVPR}, 2019, pp.
  8660--8669.

\bibitem{shi2015convolutional}
X.~Shi, Z.~Chen, H.~Wang, D.-Y. Yeung, W.-K. Wong, and W.-c. Woo,
  ``Convolutional lstm network: A machine learning approach for precipitation
  nowcasting,'' \emph{NIPS}, vol.~28, 2015.

\bibitem{tran2015learning}
D.~Tran, L.~Bourdev, R.~Fergus, L.~Torresani, and M.~Paluri, ``Learning
  spatiotemporal features with 3d convolutional networks,'' in \emph{ICCV},
  2015, pp. 4489--4497.

\bibitem{huang2019batching}
Y.~Huang, Z.~Tang, D.~Chen, K.~Su, and C.~Chen, ``Batching soft iou for
  training semantic segmentation networks,'' \emph{{IEEE} Signal Processing
  Lett.}, vol.~27, pp. 66--70, 2019.

\end{thebibliography}

\end{document}